\documentclass{article}
\usepackage{spconf,amsmath,amssymb,amsfonts,amsthm,graphicx}
\usepackage{makecell}
\usepackage{float}

\def\AE{{\textnormal{AE}}}
\title{On The Utility of Conditional Generation Based Mutual Information for Characterizing Adversarial Subspaces}
\name{Chia-Yi Hsu$^*$, Pei-Hsuan Lu$^*$, Pin-Yu Chen$^\dagger$ and Chia-Mu Yu$^*$}
\address{$^*$National Chung Hsing University, Taiwan\\
$^\dagger$IBM Research}
\begin{document}
%
\maketitle
\begin{abstract}
Recent studies have found that deep learning systems are vulnerable to \emph{adversarial examples}; e.g., visually unrecognizable adversarial images can easily be crafted to result in misclassification. The robustness of neural networks has been studied extensively in the context of adversary detection, which compares a metric that exhibits strong discriminate power between natural and adversarial examples.
In this paper, we propose to characterize the  adversarial subspaces through the lens of mutual information (MI) approximated by conditional generation methods. We use MI as an information-theoretic metric to strengthen existing defenses and improve the performance of adversary detection. Experimental results on MagNet defense demonstrate that our proposed MI detector can strengthen its robustness against powerful  adversarial attacks.
\end{abstract}
\begin{keywords}
Adversarial example, conditional generation, detection, mutual information
\end{keywords}
\section{Introduction}
\label{sec:intro}
In recent years, deep learning has demonstrated impressive performance on many tasks in machine learning, such as speech recognition and image classification. However, recent research has shown that well-trained deep neural networks (DNNs) are rather vulnerable to adversarial examples \cite{szegedy2013intriguing,goodfellow2014explaining,biggio2017wild,su2018robustness}. There have been many efforts on defending against adversarial examples. In order to enhance the robustness of DNNs against adversarial perturbations, several studies aim to characterize adversarial subspaces and develop the countermeasures. For example, Ma et al. \cite{ma2018characterizing} characterize the dimensional properties of adversarial regions through the use of local intrinsic dimensionality (LID). Nonetheless, very recently Lu et al. \cite{lu2018limitationLID} demonstrate the limitation of LID. Generally speaking, the essence of adversary detection lies in finding a metric that exhibits strong discriminative power between natural and adversarial examples. More importantly, when mounting the detector to identify adversarial inputs, one should ensure minimal performance degradation on the natural (clean) examples, which suggests a potential  trade-off between test accuracy and adversary detectability.

In this paper, we propose to characterize adversarial subspaces by using mutual information (MI). 
Our approach is novel in the sense that the MI is approximated by a well-trained conditional generator owing to the recent advances in generative adversarial networks (GANs) \cite{goodfellow2014generative}. We demonstrate the effectiveness of our approach on MagNet \cite{meng2017magnet}, a recent defense method based on data reformation and adversary detection.
Experimental results show that when integrating   MagNet with our MI detector,  the detection capability can be significantly improved against powerful adversarial attacks.

\section{Background}
\label{sec:format}

Adversarial examples can be categorized into targeted and untargeted attacks based on attack objectives. The former falsely renders the prediction of the targeted DNN model towards a specific output, while the latter simply leads the targeted DNN model to a falsified prediction. 

\subsection{Carlini and Wagner's Attack (C\&W attack)}\label{sec:Carlini and Wagner's Attack}
Carlini and Wagner \cite{carlini2017towards} propose an optimization-based framework for
targeted and untargeted attacks that can generate adversarial examples with a small perturbation. They design an $L_{2}$ norm regularized loss function in addition to the model prediction loss defined by the logit layer representations in DNNs.  C\&W attack can successfully bypass undefended and several defended DNNs by simply tuning the confidence parameter $\kappa$ in the optimization process of generating adversarial examples \cite{carlini2017adversarial}. It can also be adopted to generate adversarial examples based on $L_{0}$ and $L_{\infty}$ distortion metrics.

C\&W attack finds an effective adversarial perturbation by solving the following optimization problem:
\[\mathop{{\rm minimize}}\limits_{\delta \in \mathbb{R}^d}\quad \Vert \delta \Vert_{2}^2 + c \cdot f(x_0 + \delta) ~~{\rm such\,that}\quad x_0 +\delta \in [0,1]^{d},\]
where $d$ is the data dimension and $[0,1]^d$ denotes the space of valid data examples. 
For a natural example $x_0$, the C\&W attack aims to find a small perturbation $\delta$ (evaluated by $\Vert \delta \Vert_2$) in order to preserve the visual similarity to $x_0$ but will  simultaneously deceive the classifier (evaluated by the $f(\cdot)$ term). The hyperparameter $c$ is used to balance these two losses. 

Let $x=x_0+\delta$ denote the perturbed example of $x_0$. The loss $f$ is designed in a way that $f(x)\leq\,0$ if and only if the classifier assigns $x$ to a wrong class. In particular, for untargeted attacks $f(x)$ takes a hinge loss form defined as
\begin{align}
f(x) = \max \{ {\rm Z}(x)_{l_{x_0}} - \max( {\rm Z} (x)_{i: i \neq  l_{x_0}} ), -\kappa\},
\end{align}
where ${\rm Z}(x)$ is the hidden representation of $x$ in the pre-softmax layer (also known as the logits) and $l_{x_0}$ is the ground truth label of $x_0$. Similar loss can be defined for targeted attacks.
The parameter $\kappa \geq 0$ is a hyper-parameter governing the model confidence of $x$. Setting a  higher $\kappa$ gives a stronger adversarial example in classification confidence. 

\subsection{EAD: Elastic-Net Attack to Deep Neural Network}\label{sec:EAD: Elastic-Net Attack}
\indent Chen et al. \cite{chen2017ead} propose EAD attack which has two decision rules: one is elastic-net (EN) rule and the other is $L_{1}$ rule. In the process of attack optimization, the EN decision rule selects the minimally-distorted adversarial example based on the elastic-net loss of all successful adversarial examples in the attack iterations. On the other hand, the $L_{1}$ decision rule refers to selecting the final adversarial examples based on the minimum of $L_{1}$ distortion among successful adversarial examples. In essence, EAD attack finds an effective adversarial example by solving the following optimization problem:
\begin{eqnarray}\nonumber
&&{\rm minimize_{x \in \mathbb{R}^d}} \quad c \cdot \emph{f}({\rm x})+ \Vert {\rm x} - {\rm x}_{0}\Vert^{2}_{2} + \beta \Vert {\rm x}-{\rm x}_{0}\Vert_{1} \\
&&{\rm subject \; to}\quad {\mathrm x} \in [0,1]^d. \nonumber
\end{eqnarray}
The parameters \emph{c}, $\beta\; \geq 0$ are regularization parameters for \emph{f} and $L_{1}$ distortion. Notably, the attack formulation of C\&W attack can be viewed as a special case of EAD attack when the $L_{1}$ penalty coefficient $\beta=0$, reducing to a pure $L_{2}$ distortion based attack. In many cases, EAD attack can generate more effective  adversarial examples than C\&W attack by considering the additional $L_1$ regularization \cite{chen2017ead,sharma2017attacking,lu2018limitationLID,lu2018limitation}.

\subsection{MagNet: Defending Adversarial Examples with Detector and Reformer}\label{sec: MagNet}
Recently, strategies such as feature squeezing \cite{xu2017feature}, manifold projection \cite{meng2017magnet}, gradient and representation masking \cite{papernot2016distillation,bradshaw2017adversarial}, and adversarial training \cite{madry2017towards}, have been proposed as potential defenses against adversarial examples. 

In particular, the MagNet proposed in \cite{meng2017magnet}, which is composed of an adversary detector and a data reformer, can not only filter out adversarial examples but also rectify adversarial examples via manifold projection learned from an auto-encoder. The \emph{detector} compares the statistical difference between an input example and the reconstructed one via an auto-encoder. The \emph{reformer} trained by an auto-encoder reforms the input example and brings it close to the data manifold of training data. MagNet declares a robust defense performance against C\&W attack under different confidence levels in the \emph{oblivious} attack setting, where an attacker knows the model parameters but is unaware of MagNet defense.
 In addition, MagNet can also defend against many attacks such as DeepFool \cite{moosavi2016deepfool}, the fast gradient sign method (FGSM) \cite{goodfellow2014explaining}, and iterative FGSM \cite{kurakin2016adversarial_ICLR}. However, Lu et al. \cite{lu2018limitation} demonstrate that despite its success in defending against $L_{2}$ distortion based adversarial examples on MNIST and CIFAR-10, in the same oblivious attack setting MagNet is less effective against $L_{1}$ distortion based adversarial examples crafted by EAD attack.

\subsection{Conditional Generation}\label{sec:Conditional generation}

Generative adversarial networks (GANs) \cite{goodfellow2014generative} have been recently proposed to generate (fake) data examples that are distributionally similar to real ones from a low-dimensional latent code space in an unsupervised manner. Conditional generation, i.e., generating data examples with specific properties, 
has also been made possible by incorporating side information such as class labels into the GAN training. For example, the $\alpha$-GAN \cite{lutz2018alphagan} combines the variational auto-encoder (VAE) and a GAN for generation. The
use of VAE enables the capability of inferring the latent variable from training data in addition to the realistic generative power of GAN. 

\begin{figure}[t]
\begin{center}
\includegraphics[width=240pt,height=45.5pt]{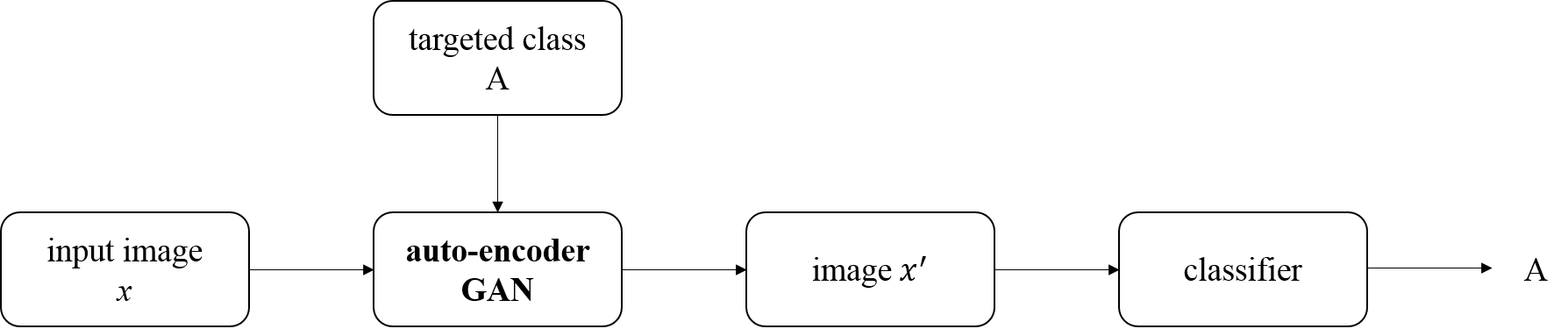}
\caption{Pipeline of conditional generation and classification.}
\label{Fig_cond_gen}
\end{center}
\vspace{-6mm}
\end{figure}

\begin{figure}[t]
\begin{center}
\includegraphics[width=240pt,height=55pt]{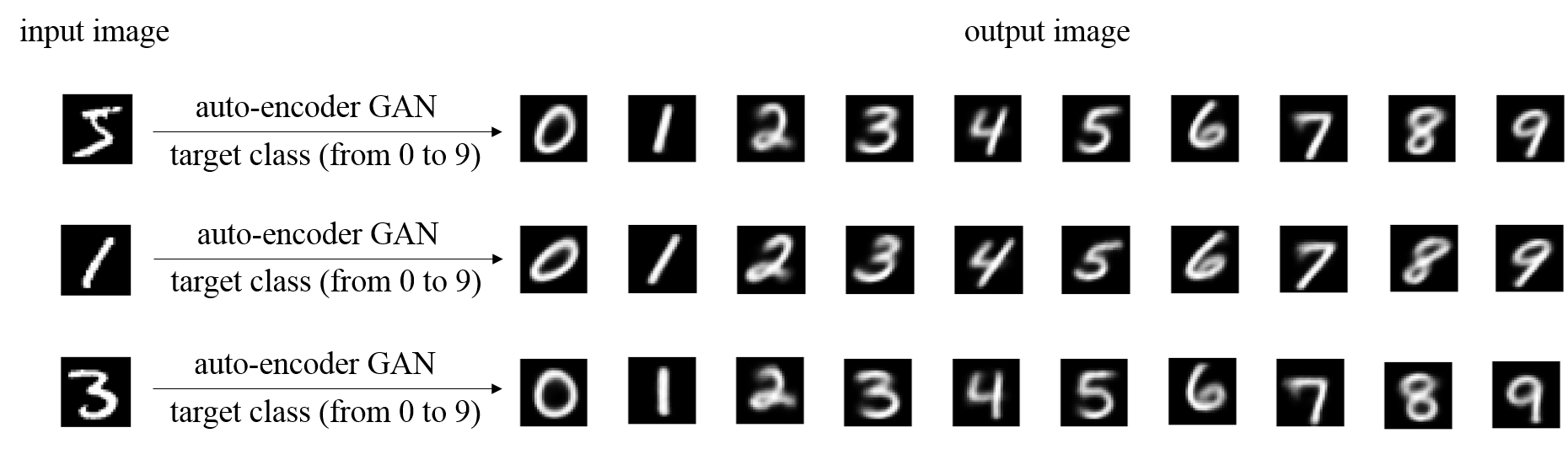}
\caption{Generated MNIST images using auto-encoder GAN given an input image and different class (digit) conditions.}
\label{Fig_cond_gen_mnist}
\end{center}
\vspace{-6mm}
\end{figure}

For the purpose of adversary detection via  mutual information approximation based on conditional generation, in this paper we adopt the 
$\alpha$-GAN framework to train a conditional generator using an auto-encoder + GAN architecture, where the class condition is appended to the generation process, as illustrated in Figure \ref{Fig_cond_gen}. Figure \ref{Fig_cond_gen_mnist} shows some generated hand-written digits of our conditional generator trained on MNIST given an input image and different class (digit) conditions.

\section{Proposed Method}
\label{sec:Our Proposed Method}
Characterizing adversarial subspaces aids in understanding the behaviors of adversarial examples and potentially gaining discriminate power against them \cite{ma2018characterizing}. In this paper, we
use an anto-encoder to learn the low-dimensional data manifold via reconstruction and propose to use it for conditional generation and approximating the  mutual information (MI) as a discriminative metric for detecting adversarial inputs. By treating an input example as an instance drawn from an oracle data generation process, our main idea roots in the hypothesis that the MI of a natural input before and after projecting to the (natural) data manifold should be maximally preserved, while the MI of an adversarial input should be relatively small due to its deviation to the data manifold. Therefore, the MI could be used as a detector to distinguish adversarial inputs.

For any two discrete random variables $X$ and $Y$, their MI is defined as  
\begin{align}
\label{eqn_MI}
  I(W,Y) = H(Y) - H(Y|W),
\end{align}
where $H(Y)$ is the entropy of $Y$, which is defined as
\begin{align}
H(Y) = -\sum_{y}\,p_Y(y)\log\,p_Y(y)
\end{align}
and $p_Y(y)$ is the probability of $Y=y$. The conditional entropy $H(Y|W)$ is defined as 
\begin{align}
 H(Y|W) = -\sum_{w} p_W(w)\left(\sum_{y}\,p_{Y|W}(y|w)\,\log\,p_{Y|W}(y|w)\right),
\end{align}
where $p_{Y|W}(y|w)$ is the probability of $Y=y$ conditioned on $W=w$.

 Connecting the dots between MI and our adversary detection propoal, let $f(\cdot):\mathbb{R}^d \mapsto \mathbb{R}^K$ denote the (DNN) classifier that takes a $d$-dimensional vector input and outputs the prediction results (i.e., probability distributions) over $K$ classes,  and let $\AE(\cdot):\mathbb{R}^d \mapsto \mathbb{R}^d$ denote the auto-encoder trained for reconstruction using training data. For any data input $X$ (either natural or adversarial), we proposes to use the MI of $f(X)$ and $f(\AE(X))$ as the metric for adversary detection. Specifically, we 
consider the setting where $W = f(X)$ and $Y = f(\AE(X))$ in \eqref{eqn_MI}, and we use the Jaccard distance (a properly normalized index between $[0,1]$)
\begin{align}
J(W,Y) = 1 - \frac {d(W,Y)}{H(W,Y)}
\end{align}
for detection, where $d(W,Y) = H(W) + H(Y)  - 2I(W,Y)$ and $H(W,Y) = H(Y) + H(W) - I(W,Y)$.
Here the Jaccard distance measures the information-theoretic difference between $W$ and $Y$. In the adversary detection setting, large 
Jaccard distance means that $f(X)$ and $f(\AE(X))$
are more distinct in distribution and thus indiate $X$ and $\AE(X)$ share less similarity. Therefore, we declare $X$ as an adversarial input if $J(f(X),f(\AE(X))) \geq \eta$, where $\eta$ is a pre-specified threshold that balances adversary detectability and rejection rate on natural inputs.

We note that while the entropy $H(f(\AE(X)))$ can be easily computed, the conditional entropy $H(f(\AE(X))|f(X))$ is difficult to be evaluated when $f$ is a DNN classifier. 
Here we propose to use the conditional generator as introduced in Section \ref{sec:Conditional generation} to approximate the conditional entropy.
In particular, the conditional probability $p_{f(\AE(X))|f(X)}(y|w)$ is evaluated by the prediction probability of the generated image $X^\prime$ to be classified as $y$ given the class condition $w$ and the latent code of $X$ as the input to the conditional generator.

\begin{table*}[h]
\caption{Comparison of MagNet and MI-strengthened MagNet  on MNIST in terms of classification accuracy (\%) of adversarial examples. Our MI detector can improve the detection performance by up to 31.2\%.}
\label{Table I}
\resizebox{170mm}{30mm}{
\begin{tabular}{c|ccc|lll}
\hline 
\multicolumn{7}{c}{MNIST} \\ 
\hline 
•&\multicolumn{3}{c}{MagNet} & \multicolumn{3}{c}{MI-strengthened MagNet}  \\ 
\hline
\makecell{Attack \\method}& \makecell{C\&W attack\\($L_{2}$ version)} & \makecell{EAD attack\\($L_{1}$ rule, $\beta = 10^{-1}$)} & \makecell{EAD attack\\(EN rule, $\beta = 10^{-1}$)}&  \makecell{C\&W attack\\($L_{2}$ version)} & \makecell{EAD attack\\($L_{1}$ rule, $\beta = 10^{-1}$)}  & \makecell{EAD attack\\(EN rule, $\beta = 10^{-1}$)}\\ 
\hline 
$\kappa$ &• & • & • & • & • & • \\ 
0 & 98.7 & 78.8 & 78.1 & \hspace{0.2cm}98.7 & \hspace{1cm}78.8 & \hspace{1cm}78.1 \\  
5 & 94.6 & 33.5 & 26.6 & \hspace{0.2cm}95.8\;($\uparrow1.2$) & \hspace{1cm}39.4\;($\uparrow5.9$) & \hspace{1cm}37.4\;($\uparrow10.8$)\\    
10 & 91.5 & 17.9 & 11.7 & \hspace{0.2cm}97.8\;($\uparrow6.3$) & \hspace{1cm}46.9\;($\uparrow29$) & \hspace{1cm}44.0\;($\uparrow32.3$) \\ 
15 & 90.0 & 16.2 & 9.7 & \hspace{0.2cm}98.0\;($\uparrow8.0$) & \hspace{1cm}47.4\;($\uparrow31.2$) & \hspace{1cm}41.8\;($\uparrow32.1$) \\ 
20 & 91.4 & 19.6 & 12.1 & \hspace{0.2cm}98.2\;($\uparrow6.8$) & \hspace{1cm}45.1\;($\uparrow25.5$) & \hspace{1cm}36.8\;($\uparrow24.7$) \\ 
25 & 93.9 & 26.1 & 16.8 & \hspace{0.2cm}98.4\;($\uparrow4.5$) & \hspace{1cm}44.3\;($\uparrow18.2$) & \hspace{1cm}35.6\;($\uparrow18.8$) \\ 
30 & 96.2 & 34.5 & 22.5 & \hspace{0.2cm}98.5\;($\uparrow2.3$) & \hspace{1cm}44.3\;($\uparrow9.8$)& \hspace{1cm}32.9\;($\uparrow10.4$) \\ 
35 & 97.7 & 41.1 & 28.6 & \hspace{0.2cm}99.0\;($\uparrow1.3$) & \hspace{1cm}47.3\;($\uparrow6.2$) & \hspace{1cm}35.4\;($\uparrow6.8$) \\ 
40 & 98.5 & 47.8 & 33.1 & \hspace{0.2cm}98.9\;($\uparrow0.4$) & \hspace{1cm}52.0\;($\uparrow4.2$) & \hspace{1cm}37.9\;($\uparrow4.8$)\\
\hline
\end{tabular}}
\vspace{-4mm}
\end{table*}

\begin{table*}[h]
\caption{Comparison of MagNet and MI-strengthened MagNet  on CIFAR-10 in terms of classification accuracy (\%) of adversarial examples. Our MI detector can improve the detection performance by up to 3.8\%.}
\label{Table II}
\resizebox{170mm}{30mm}{
\begin{tabular}{c|ccc|lll}
\hline 
\multicolumn{7}{c}{CIFAR-10} \\ 
\hline 
•&\multicolumn{3}{c}{MagNet} & \multicolumn{3}{c}{MI-strengthened MagNet}  \\ 
\hline
\makecell{Attack \\method}& \makecell{C\&W attack\\($L_{2}$ version)} & \makecell{EAD attack\\($L_{1}$ rule, $\beta = 10^{-1}$)} & \makecell{EAD attack\\(EN rule, $\beta = 10^{-1}$)}&  \makecell{C\&W attack\\($L_{2}$ version)} & \makecell{EAD attack\\($L_{1}$ rule, $\beta = 10^{-1}$)}  & \makecell{EAD attack\\(EN rule, $\beta = 10^{-1}$)}\\ 
\hline 
$\kappa$ &• & • & • & • & • & • \\ 
0 & 80.1 & 70.5 & 70.7 & \hspace{0.2cm}80.1 & \hspace{1cm}70.3 & \hspace{1cm}70.6 \\  
10 & 50.3 & 26.2 & 26.4& \hspace{0.2cm}51.3\;($\uparrow1.0$) & \hspace{1cm}28.4\;($\uparrow2.2$) & \hspace{1cm}29.4\;($\uparrow3.0$) \\    
20 & 48.0 & 26.8& 26.8 & \hspace{0.2cm}51.8\;($\uparrow3.8$) & \hspace{1cm}29.4\;($\uparrow2.6$) & \hspace{1cm}29.1\;($\uparrow2.3$) \\ 
30 & 62.9 & 37.1 & 38.4 & \hspace{0.2cm}64.0\;($\uparrow1.1$) & \hspace{1cm}38.6\;($\uparrow1.5$)& \hspace{1cm}39.4\;($\uparrow1.0$) \\ 
40 & 72.3 & 48.4 & 45.3 & \hspace{0.2cm}73.0\;($\uparrow0.7$) & \hspace{1cm}49.3\;($\uparrow0.9$) & \hspace{1cm}46.7\;($\uparrow1.4$) \\ 
50 & 81.4 & 61.0 & 60.0 & \hspace{0.2cm}81.8\;($\uparrow0.4$) & \hspace{1cm}61.2\;($\uparrow0.2$) & \hspace{1cm}60.3\;($\uparrow0.3$) \\ 
60 & 89.6 & 73.8 & 71.7 &\hspace{0.2cm}89.6 & \hspace{1cm}74.0\;($\uparrow0.2$) & \hspace{1cm}71.7 \\ 
70 & 94.6 & 84.6 & 81.5 & \hspace{0.2cm}94.6 & \hspace{1cm}84.6 & \hspace{1cm}81.6\;($\uparrow0.1$) \\ 
80 & 97.3 & 90.6 & 90.4 & \hspace{0.2cm}97.3 & \hspace{1cm}90.7\;($\uparrow0.1$) & \hspace{1cm}90.4\\
\hline
\end{tabular}}
\end{table*}
\section{Experiments}
\label{sec:Experiments}

In this section, we applied the proposed MI detector to the MagNet defense
against untargeted C\&W and EAD attacks on MNIST and CIFAR-10 datasets under the oblivious attack setting. 
For each dataset, we randomly selected 1000 correctly classified images from the test sets to generate adversarial examples with different confidence levels. 
Under this attack setting, it has been shown in \cite{lu2018limitation} that while MagNet is resilient to C\&W attack, it is more vulnerable to EAD attack.

\subsection{Experiment Setup and Parameter Setting}\label{sec:Experiment Setup and Parameter Setting}
 We followed the oblivious attack setting used in MagNet\footnote{https://github.com/Trevillie/MagNet}, where the adversarial examples are generated from the same DNN but the adversary is unaware of the deployed defense. The image classifiers on MNIST and CIFAR-10 are trained with the same DNN architecture and training parameters in \cite{meng2017magnet}. 
 For defending adversarial inputs, we report the classification accuracy measured by the percentage of adversarial examples detected by the detectors or correctly classified after passing the reformer. Higher classification accuracy means better defense performance.

Similar to MagNet, in each dataset a validation set of 1000 images is used to determine the MI detector threshold $\eta$, which is set such that the false-positive rate is 0.5\%. The $\alpha$-GAN framework \cite{lutz2018alphagan} was used to train our conditional generator consisting of an auto-encoder + GAN architecture.



We report the defense results of untargeted attacks since they are generally more difficult to be detected than targeted attacks. For C\&W attack\footnote{https://github.com/carlini/nn robust attacks.} ($L_{2}$ version), we used the same parameters in \cite{meng2017magnet} to generate adversarial examples. For EAD attack\footnote{https://github.com/ysharma1126/EAD-Attack}, we used the same parameters in \cite{lu2018limitation} and generate adversarial examples using both elastic-net (EN) and $L_{1}$
decision rules. All experiments were conducted using an Intel Xeon E5-2620v4 CPU, 125 GB RAM and a NVIDIA TITAN Xp GPU with 12 GB RAM.


\subsection{Performance Evaluation on MNIST}\label{sec:Evaluation of Black-Box Transfer Attacks to MI on MNIST}

\noindent \textbf{Effect on natural examples.} Without MagNet, the original test accuracy is 99.42\%. With MagNet, it is decreased to 99.13\%; with  MI-strengthened MagNet, the test accuracy could still remain at 98.63\%.
The slight reduction in test accuracy trades in enhanced adversary detectability.

\noindent \textbf{Effect on adversarial examples.} Table \ref{Table I} shows the classification accuracy of adversarial examples on MNIST. While MagNet is robust to C\&W attack in the oblivious attack setting, it is shown to be less effective against EAD attack \cite{lu2018limitation}, especially for medium confidence levels (e.g., $\kappa=\{10,15,20\}$). On the other hand, our MI-strengthened MagNet can improve the detection performance by up to 31.2\%, highlighting the utility of approximating MI via conditional generation for characterizing adversarial subspaces.


\subsection{Performance Evaluation  on CIFAR-10}\label{Evaluation of Black-Box Transfer Attacks to MI on CIFAR-10}
\noindent \textbf{Effect on natural examples.} Without MagNet,  the test accuracy is 86.91\%. With MagNet, it reduced to 83.33\%; with  MI-strengthened MagNet, it becomes 83.18\%. 

\noindent \textbf{Effect on adversarial examples.}
Comparing to MNIST, on CIFAR-10 the MI-strengthened MagNet provides less improvement (up to 3.8\%) in adversary detection. One possible explanation is that for most $\kappa$ values the original MagNet already performs better on CIFAR-10 than on MNIST. We also conjecture that the performance of MI detector is closely associated with the capability of the conditional generator, as we observe that the quality of the generated images on MNIST is much better than those on CIFAR-10.

\section{Conclusion}
\label{sec:Conclusion}
In this paper, we propose to utilize the mutual information (MI) and data manifold of deep neural networks as a novel information-theoretic metric to characterize and distinguish adversarial inputs, where the MI is approximated by a well-trained conditional generator. The experimental results show that our MI detector can effectively strengthen the detection capability of MagNet defense while causing negligible effect on test accuracy. Our future work includes further exploring the utility of MI and conditional generation for adversarial robustness and extending our detector to other defenses.


\bibliographystyle{IEEEbib}
\bibliography{adversarial_learning}

\end{document}